\documentclass[final]{cvpr}

\usepackage{times}
\usepackage{epsfig}
\usepackage{graphicx}
\usepackage{amsmath}
\usepackage{amssymb}
\usepackage[utf8]{inputenc}

\usepackage[pagebackref=true,breaklinks=true,colorlinks,bookmarks=false]{hyperref}

\def\cvprPaperID{****} 
\def\confYear{CVPR 2021}

\begin{document}

\title{Continual Learning on the Edge with TensorFlow Lite}

\author{Giorgos Demosthenous\\
CYENS Centre of Excellence\\
Dimarchias Square 23, Nicosia 1016, Cyprus\\
{\tt\small g.demosthenous@cyens.org.cy}
\and
Vassilis Vassiliades \\
CYENS Centre of Excellence\\
Dimarchias Square 23, Nicosia 1016, Cyprus\\
{\tt\small v.vassiliades@cyens.org.cy}
}

\maketitle

\begin{abstract}
   Deploying sophisticated deep learning models on embedded devices with the purpose of solving real-world problems is a struggle using today’s technology. Privacy and data limitations, network connection issues, and the need for fast model adaptation are some of the challenges that constitute today’s approaches unfit for many applications on the edge and make real-time on-device training a necessity. Google is currently working on tackling these challenges by embedding an experimental transfer learning API to their TensorFlow Lite, machine learning library. In this paper, we show that although transfer learning is a good first step for on-device model training, it suffers from catastrophic forgetting when faced with more realistic scenarios. We present this issue by testing a simple transfer learning model on the CORe50 benchmark as well as by demonstrating its limitations directly on an Android application we developed. In addition, we expand the TensorFlow Lite library to include continual learning capabilities, by integrating a simple replay approach into the head of the current transfer learning model. We test our continual learning model on the CORe50 benchmark to show that it tackles catastrophic forgetting, and we demonstrate its ability to continually learn, even under non-ideal conditions, using the application we developed. Finally, we open-source the code of our Android application to enable developers to integrate continual learning to their own smartphone applications, as well as to facilitate further development of continual learning functionality into the TensorFlow Lite environment.
\end{abstract}

\section{Introduction}

In recent years, there has been a rising trend of taking sophisticated deep learning models out of the lab and deploying them on the edge (e.g., smartphones, robotics and other embedded devices). There has also been an ongoing effort to discover real-world, everyday use, through mainstream adoption of this technology. Most approaches today, focus on training their machine learning models offline, on powerful GPU-supported servers, and deploying them on the edge for inference. Although this approach is good enough for a variety of deep learning applications, many times it is desirable to allow on-device training of the models as well. Privacy limitations, network connection issues, personalization and fast adaptation are some of the reasons why online, on-the-edge training, is necessary. Google, who backs the development of the TensorFlow (TF) and TensorFlow Lite machine learning platforms, has picked up on this growing need and is now working on embedding on-device training functionality to their TensorFlow Lite library. Their current approach focuses on using transfer learning for on-device personalization which they introduce through a pre-release API~\cite{senchanka_2019}.

Transfer Learning (TL) has been shown to perform extremely well when faced with learning new classes by training on just a few samples from each class~\cite{pan2009survey}. When deployed, it allows the model to learn fast and online without the need of transferring data back to the server. In order for transfer learning to work though, all new classes must be present at the training batch while learning, otherwise the model will suffer from catastrophic forgetting~\cite{mccloskey1989catastrophic} when new classes appear incrementally, over time. Unfortunately, in most real-world scenarios, new classes do appear incrementally which constitutes transfer learning unfit for many specialized applications on embedded devices. Continual Learning (CL), which is the ability to continually learn from incrementally appearing information, has been gaining traction as a potential solution to catastrophic forgetting. There is a plethora of studies around different continual learning approaches~\cite{hadsell2020embracing,parisi2019continual}, as well as methods for benchmarking and evaluating continual learning models~\cite{hsu2018re,farquhar2018towards,van2019three,lomonaco2017core50}.
Paradoxically and to the best of our knowledge, there is only a handful of studies that focus on deploying continual learning on embedded devices~\cite{li2019rilod, pellegrini2019latent, doshi2020continual}, and no study, in particular, on doing so on a mobile phone using TF Lite.


The contributions of this paper are as follows:
\begin{itemize}
   \item We expand the current TF Lite capabilities by integrating continual learning into its API, specifically by introducing a simple replay buffer.
   \item We experimentally compare models trained using TL and CL under various scenarios on an Android application we developed, demonstrating both the necessity of CL and its superiority over TL in realistic settings.
   \item We open-source the code of our Android application to facilitate further development of continual learning functionality within the TF Lite environment.
\end{itemize}


In Section~\ref{sec3}, we describe how the TL and CL models are built using TF Lite and examine how they perform under the CORe50 benchmark~\cite{lomonaco2017core50,lomonaco2019rehearsal}. In addition, we compare these models with AR1*~\cite{pellegrini2019latent}, which is a state-of-the art continual learning algorithm. Section~\ref{sec4} demonstrates how TL and CL perform in real-world scenarios through our developed Android application.
Finally, in Section~\ref{sec5}, we draw some conclusions and we discuss how our CL model can be enhanced within TF Lite capabilities to further improve its performance and support more classes in the future.

\begin{figure}[t]
\begin{center}
\includegraphics[width=1\linewidth]{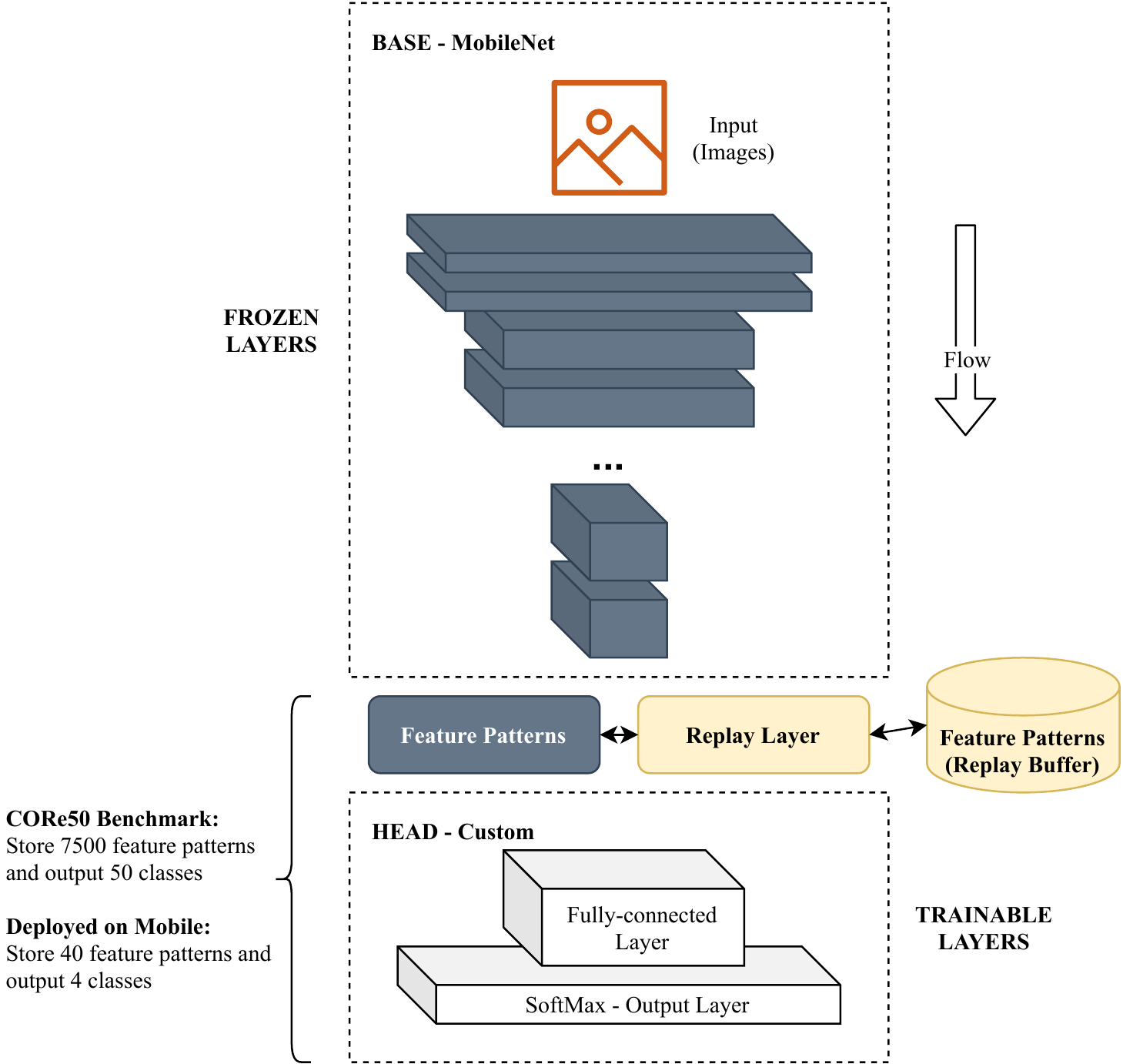}
\end{center}
   \caption{Architecture of Continual Learning model built using TensorFlow Lite.}
\label{fig:abstract}
\end{figure}

\section{Background}

The majority of the image recognition, object detection and other computer vision related tasks are tackled by training sophisticated deep neural networks, on millions of images for extended periods of time. The lack of computational resources makes it impractical to train such models in real-time while new streams of data arrive, in order to solve real-world problems. Transfer Learning, combats this issue by enabling models to use prior knowledge to solve new, similar tasks. Specifically, the model is separated into two parts, most commonly known as \emph{base} and \emph{head}. The base model is a network that was pretrained to solve a computer vision task of similar nature. In our case, a MobileNet~\cite{howard2017mobilenets} is used since we are looking to deploy the models on embedded devices and the base network is pretrained on the ImageNet dataset~\cite{deng2009imagenet}. The head network, which usually consists of a simple fully-connected layer and a softmax activation, uses the features extracted from the base to learn new classes in real time, using only a very limited amount of data. TF Lite currently offers an experimental API~\cite{senchanka_2019} that allows for extending the head network and deploying the final TL model on embedded devices. This flexibility of TF Lite makes it ideal for experimenting with continual learning capabilities and limitations, directly on the edge. The head network can be altered to facilitate lighter versions of well-known CL approaches, like architecture based methods~\cite{zhou2012online,cortes2017adanet}, regularization approaches~\cite{kirkpatrick2017overcoming,li2017learning}, memory based systems~\cite{lopez2017gradient,kemker2017fearnet} and even meta-learning~\cite{lemke2015metalearning}.

In this paper, we expand the head network by integrating a simple replay buffer as seen in Figure~\ref{fig:abstract}. Before deploying and testing the model on an Android device, we first compare it with the AR1*+Latent Replay (AR1*LR)~\cite{pellegrini2019latent} continual learning algorithm. AR1*LR is a state-of-the art continual learning algorithm that combines architectural, regularization and memory approaches to allow the entire network to continually learn. It uses a modified version of the MobileNet and is also pretrained on the ImageNet. For evaluation and model comparison purposes we use the CORe50 dataset~\cite{lomonaco2017core50} which consists of 50 different classes extracted from short videos of objects over time. Specifically, we make use of the New Instances and Classes (NICv2)~\cite{lomonaco2019rehearsal} scenario of the benchmark, since it encapsulates perfectly real-world situations where new classes are learned incrementally, by training on small, non i.i.d., batches over time. The new classes and instances of classes are presented to the model incrementally, over time in 391 batches. The accuracy of the model is evaluated after each training batch to examine its ability to continually learn.

The CORe application~\cite{pellegrini2019latent}, is an Android implementation of the AR1*LR algorithm, where the latent replay resides at the pool6 layer. It was developed using Caffe, cross-compiled for Android, and C++. It stores 500 patterns in the replay buffer and allows the user to add up to 5 new classes. When samples are captured or replayed, the entire network can be trained, which has the benefit of continually tuning the representation and have increased accuracy, but with the downside of some computational overhead. In contrast, the continual learning application we developed only trains the \emph{head} of the model, which decreases accuracy but is more time-efficient and requires less computational resources. In future work, the network included in the head will be extended to allow for tuning the representation as well. Our model currently stores just 40 patterns in the replay buffer and allows the user to add up to 4 new classes. Choosing to use TF Lite for our implementation, means that our code and model are optimized for running on embedded devices, as opposed to CORe. It also makes our project a lot easier to replicate, expand and implement in new Android applications.

\section{Enhancing TensorFlow Lite Capabilities with Continual Learning} \label{sec3}

\begin{figure}[t]
\begin{center}
\includegraphics[width=1\linewidth]{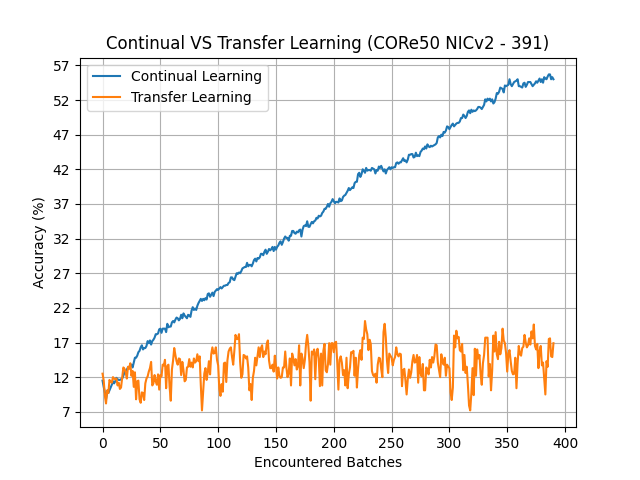}
\end{center}
   \caption{Comparing the Transfer Learning model with the Continual Learning model, in terms of accuracy over time, on the CORe50 NICv2 - 391 benchmark.}
\label{fig:cltlcomparison}
\end{figure}

Before integrating CL to TF Lite, we first need to investigate how a plain transfer learning model performs in the CORe50 NICv2 scenario. The TL model consists of a MobileNet as the base, pretrained on ImageNet, and a simple one fully-connected layer paired with a softmax activation as the head. We replace Batch Normalization layers in the MobileNet, with Batch Renormalization~\cite{ioffe2017batch} since it facilitates continual learning as shown in ~\cite{lomonaco2019rehearsal}. As depicted in Figure~\ref{fig:cltlcomparison}, transfer learning fails dramatically in an incremental learning scenario and it seems completely unable to retain old knowledge, hence the low accuracy over time, ranging between 7\% and 20\%. Therefore, it is evident that there is a lot of room for improvement when it comes to using models on embedded devices for more realistic scenarios.

We chose to implement a naïve rehearsal approach with a few modifications for our TF Lite continual learning integration. The abstract architecture of our CL model can be seen in Figure~\ref{fig:abstract}. We used the exact same model as the TL one, but with the addition of a replay buffer at the head. Instead of storing entire image samples in the buffer, we save feature patterns extracted by the base model. This way our deployed model uses a lot less resources when replaying and storing samples. For the fully-connected layer in the head we used 128 neurons, the SGD optimizer and the ReLU activation function. After each training batch during the CORe50 benchmark, the samples in the buffer are replayed to the head model. Before we move on to the next batch, we store new feature patterns into the replay buffer. If the replay buffer becomes full, we replace 1.5\% of the current samples with new ones.

\begin{figure}[t]
\begin{center}
\includegraphics[width=1\linewidth]{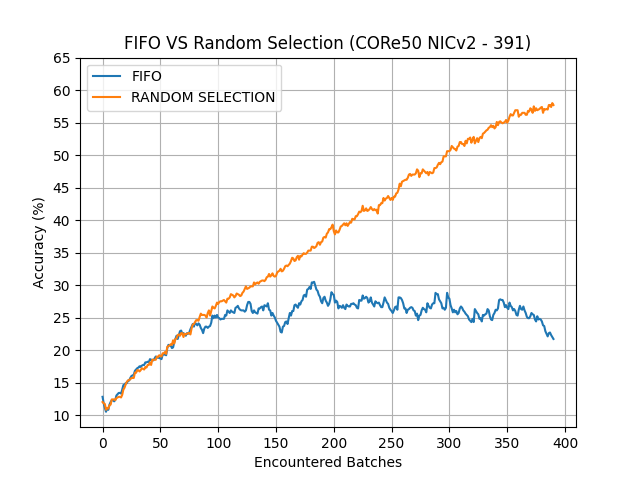}
\end{center}
   \caption{Replacing 1.5\% of old samples in the Replay Buffer with new ones, using the First-In-First-Out approach versus Random Selection and evaluating the Continual Learning model on the CORe50 NICv2 - 391 benchmark.}
\label{fig:selectionexperiments}
\end{figure}

Initially, we replaced feature patterns based on the first in first out (FIFO) principle, but the model failed to continually learn after approximately 100 training batches. This behaviour most likely occurred due to the lack of proper representation of each class in the replay buffer. Switching from FIFO to random replacement of old samples with new ones solved this issue as it can be seen in Figure~\ref{fig:selectionexperiments}. Ensuring a better class representation in the replay buffer and a more sophisticated sample replacement method will most likely further improve the performance of our CL model but it is currently outside the scope of this paper and is left for future work.

\begin{figure}[t]
\begin{center}
\includegraphics[width=1\linewidth]{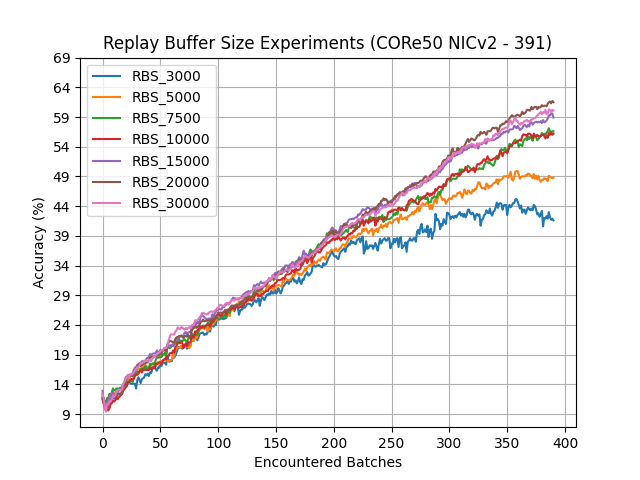}
\end{center}
   \caption{Evaluating the Continual Learning model on the CORe50 NICv2 - 391 benchmark using different maximum sample sizes for the Replay Buffer.}
\label{fig:buffersize}
\end{figure}

We conducted a series of experiments to demonstrate the trade-off between higher accuracy and replay buffer size as seen in Figure~\ref{fig:buffersize}. Having a buffer of 7500 feature patterns seems like a good balance for this benchmark scenario, since it requires 75\% less storage compared to the 30000 buffer size and the model accuracy is only reduced by 3.5\%. As depicted in Figure~\ref{fig:cltlcomparison}, integrating the aforementioned replay buffer to the head of the TL model, is a good first step to giving continual learning capabilities to the deployed TF Lite model.

We use AR1*LR as an upper bound to our model. Lowering its' latent replay buffer at the lowest possible layer of the MobileNet (pool6) makes this model comparable to ours. In that case, and as seen in Table 1, we achieve an accuracy close to that of AR1*LR, which means our model performs well-enough and is ready to be deployed on an embedded device.

\begin{table}
\begin{center}
\begin{tabular}{|l|c|}
\hline
Models & Last Accuracy \\
\hline\hline
TL & 16.9\% \\
CL & 56.6\% \\
AR1*LR & 59.76\% \\
\hline
\end{tabular}
\end{center}
\caption{The last accuracy of the Transfer Learning model, the Continual Learning model and AR1*~\cite{pellegrini2019latent} on the CORe50 NICv2 - 391 benchmark.}
\end{table}


\section{Deploying on Embedded Devices with TensorFlow Lite} \label{sec4}

In order to compare their performance in real-world scenarios, both TL and CL models were deployed on a Samsung Galaxy S10, Android device using TensorFlow Lite. Specifically, we used TensorFlow’s experimental transfer learning API as provided in their model personalization example~\cite{senchanka_2019}. We expanded both their API as well as the demo application to provide continual learning functionality and to facilitate the demonstration and comparison of the TL and CL models. Inside the application, the user can add samples for 4 different classes using the camera feed. These samples can be used to train both models in batches or incrementally, depending on what scenario the user wants to test. After training, the user can switch to TL or CL inference or even reset the models entirely. The interface of the demo application can be seen in Figure~\ref{fig:cumulativescenario}. The APK and the source code for the Android application are available on our GitLab page\footnote{\url{https://gitlab.com/riselear/public/continual-learning-on-the-edge-with-tensorflow-lite}}.

\subsection{Experimenting with Different Scenarios}

\begin{figure}[t]
\begin{center}
\includegraphics[width=1\linewidth]{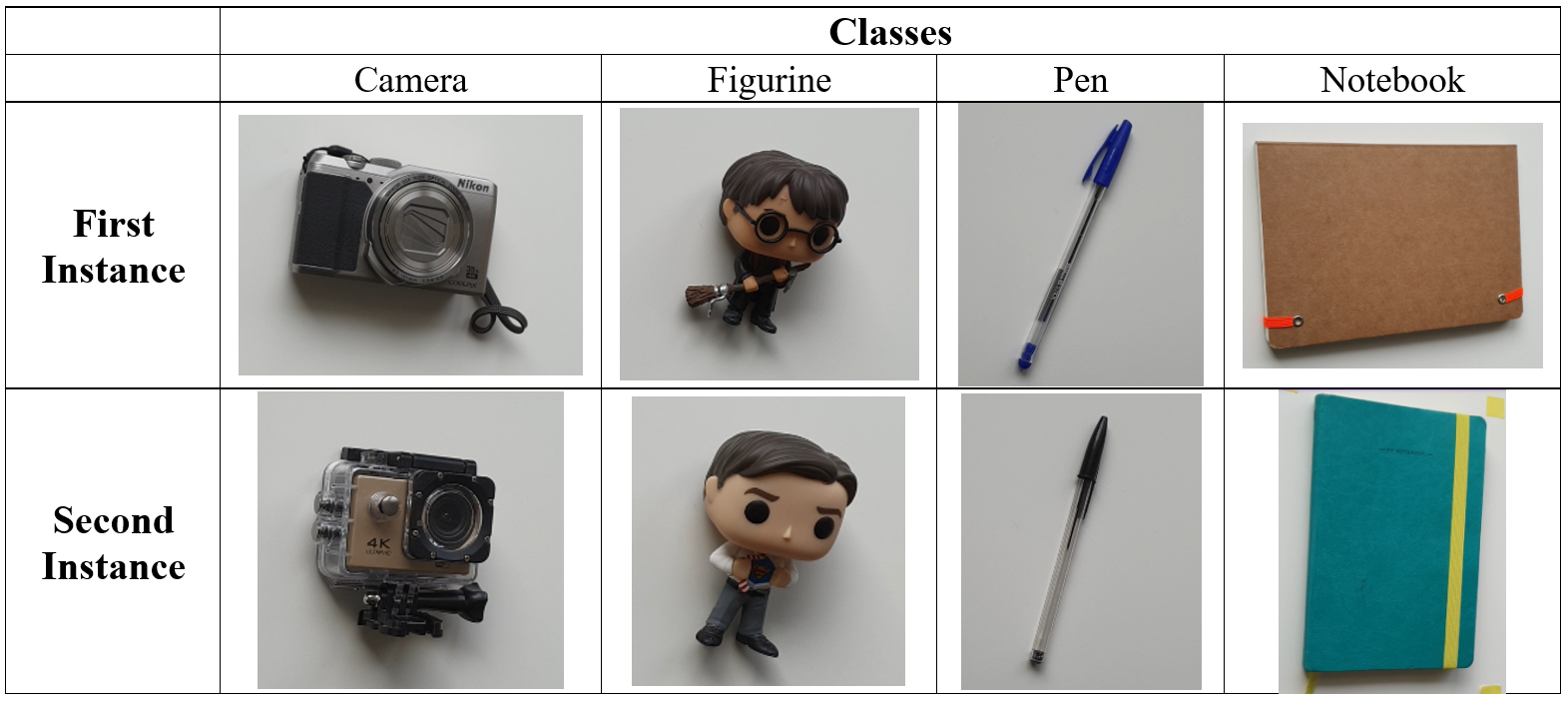}
\end{center}
   \caption{The two instances of the four classes used during the experiments on the Android demo application.}
\label{fig:classes}
\end{figure}

To demonstrate the strengths and capabilities of the CL model over the TL one, we incorporate three different scenarios in our experiments:

\begin{itemize}
  \item \textbf{Cumulative Scenario – Batch Training:} For this scenario, we add 50 samples from each of the four available classes seen in the First Instance row of Figure~\ref{fig:classes}. During sample collection, the position and rotation of the objects changes within a limited area of capture. We then use all 200 samples to train both models before evaluating their performance. In this scenario, zero samples are replayed for the CL model.
  \item \textbf{New Instances Scenario – Batch Training:} Just like with the Cumulative Scenario, we add the 50 samples per class from the First Instance row and train both models using all 200 samples. The replay buffer is populated by 10 random samples from each class. We then introduce new instances of the same four classes as seen in the Second Instance row of Figure~\ref{fig:classes}. After adding 20 new samples per class, using the new instances, we train both models with all 80 samples and replay the 40 samples from the buffer to the CL model. We evaluate the performance of the TL and CL models by their ability to remember the first instances of each class.
  \item \textbf{New Classes Scenario – Incremental Training:} The New Classes Scenario most closely resembles a natural setting and use of the deployed models. In this case, we incrementally introduce new classes to the models. In contrast with the previous scenarios, we add 50 samples for a class and train both models before moving to the next class. Every time a new class is introduced, old samples that are already in the buffer are replayed to the CL model and 10 random samples from the new class are added to the replay buffer afterwards. We repeat until both models have incrementally trained on all four classes. We then evaluate the TL and CL models by their ability to classify the objects seen in the First Instance row of Figure~\ref{fig:classes} during inference.
\end{itemize}

\begin{figure}[t]
\begin{center}
\includegraphics[width=1\linewidth]{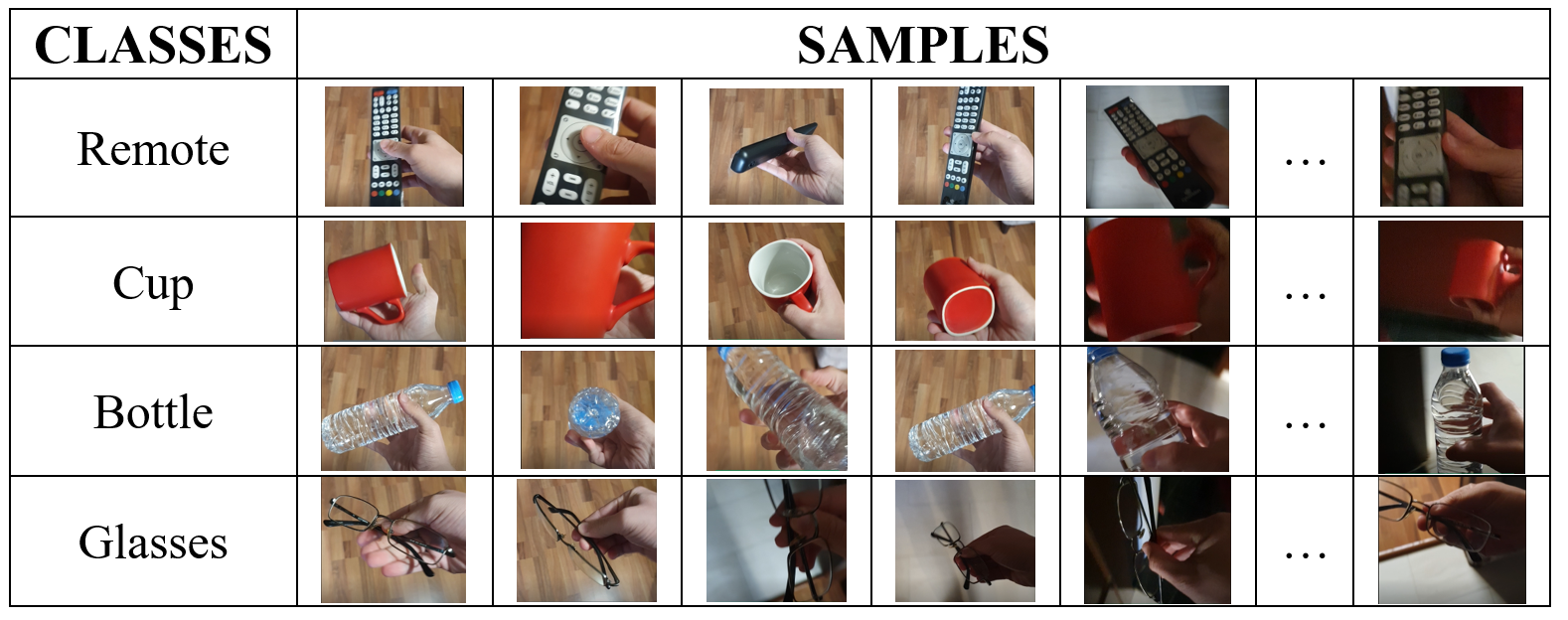}
\end{center}
   \caption{Examples of samples captured for the four classes used for training and testing the Continual Learning model, under non-ideal conditions, while deployed on an Android device.}
\label{fig:extreme}
\end{figure}

\begin{figure*}
\begin{center}
\includegraphics[width=0.95\linewidth]{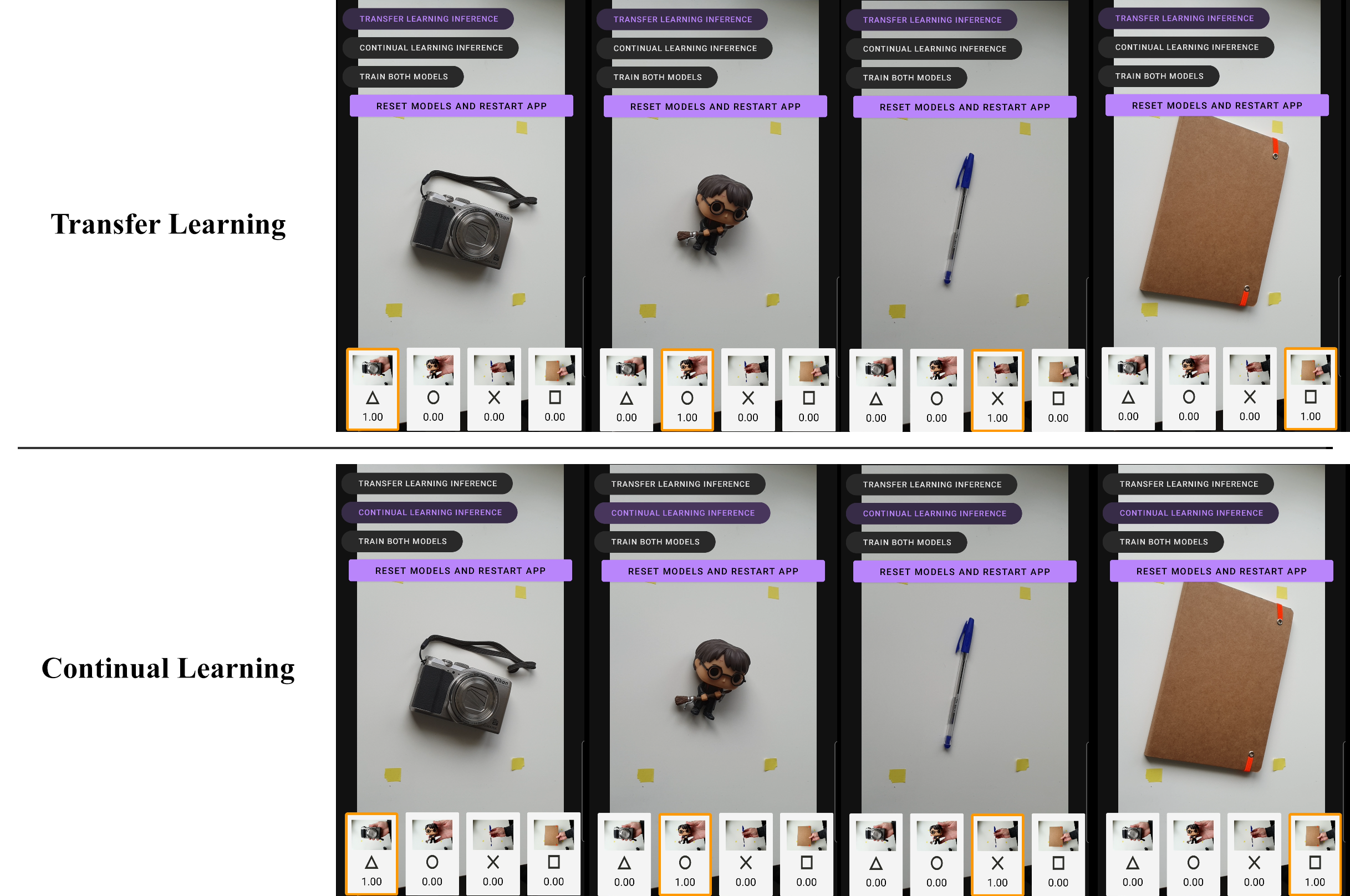}
\end{center}
   \caption{Object classification and confidence scores for both Transfer and Continual Learning models, recorded on the Android application during the Cumulative Scenario experiments. Both models have the same performance.}
\label{fig:cumulativescenario}
\end{figure*}

The entire training and evaluation procedure for each scenario can be seen in our demonstration video: \url{https://www.youtube.com/watch?v=OUvWhQouSu8}

When all data is available during training, like the Cumulative Scenario, both TL and CL models should have no issue learning and classifying new classes using only a limited number of samples. Both models should also have the same exact confidence scores during inference for each class since the replay buffer is not used in this case. Indeed, as seen in Figure~\ref{fig:cumulativescenario}, both TL and CL models correctly classify all four classes while having the same, maximum, confidence score for each class. The orange rectangle denotes which class is selected during inference and the confidence score is visible at the bottom of each class button.

\begin{figure*}
\begin{center}
\includegraphics[width=0.95\linewidth]{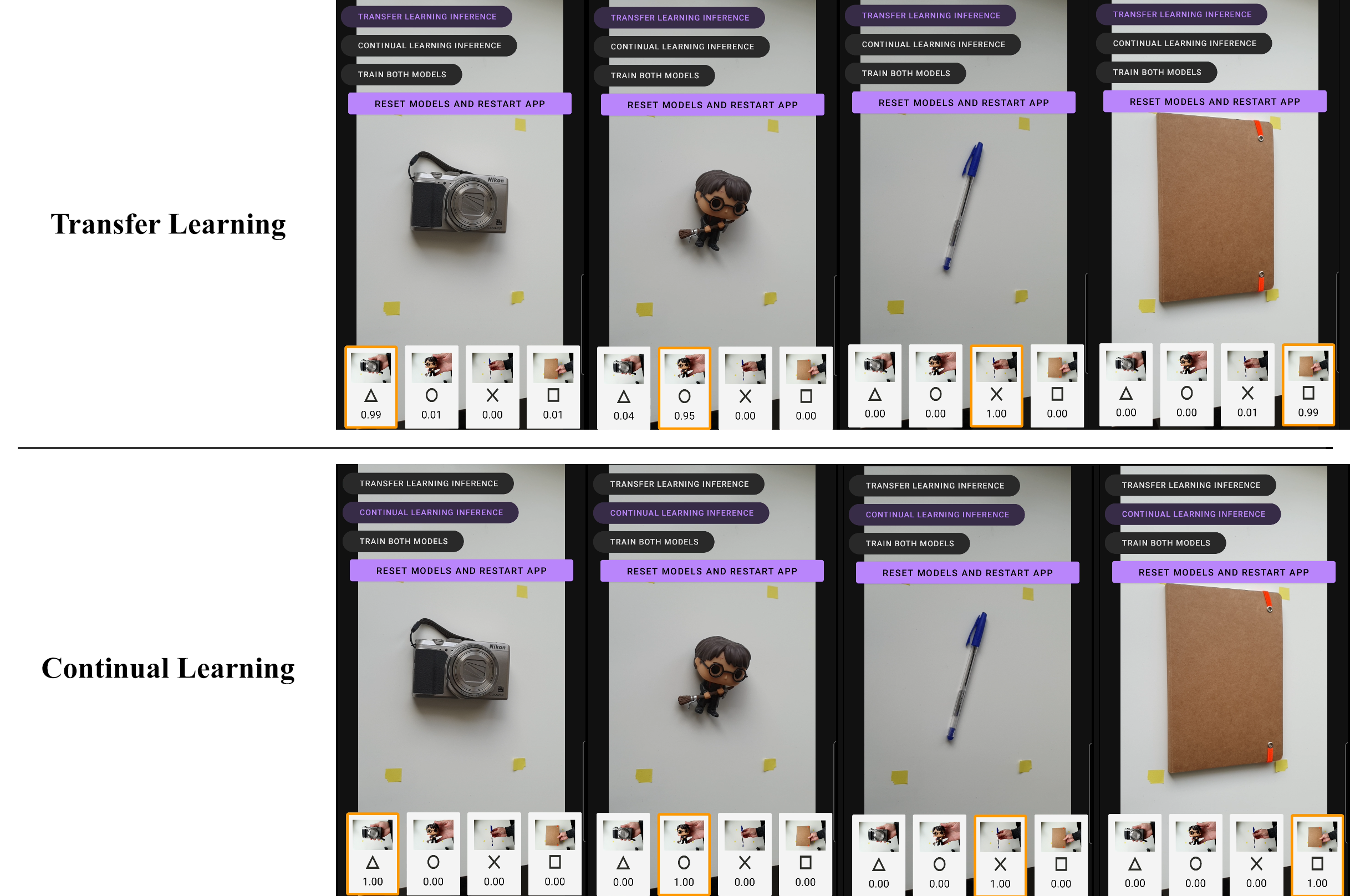}
\end{center}
   \caption{Object classification and confidence scores for both Transfer and Continual Learning models, recorded on the Android application during the New Instances Scenario experiments. Transfer Learning shows the first evidence of forgetting, even with only one new instance presented for training.}
\label{fig:instancesscenario}
\end{figure*}

In the case of the New Instances Scenario, we can see in Figure~\ref{fig:instancesscenario} that the CL model has no issue classifying correctly and with prefect scores, the first instances of the objects. TL on the other hand, despite having all new-instance data available during training, is already showing some evidence of forgetting with slightly reduced confidence scores. Although the effect of introducing only one new instance is minimal, we expect that the TL model's performance will continue to decay as new instances are introduced in the future.

\begin{figure*}
\begin{center}
\includegraphics[width=0.95\linewidth]{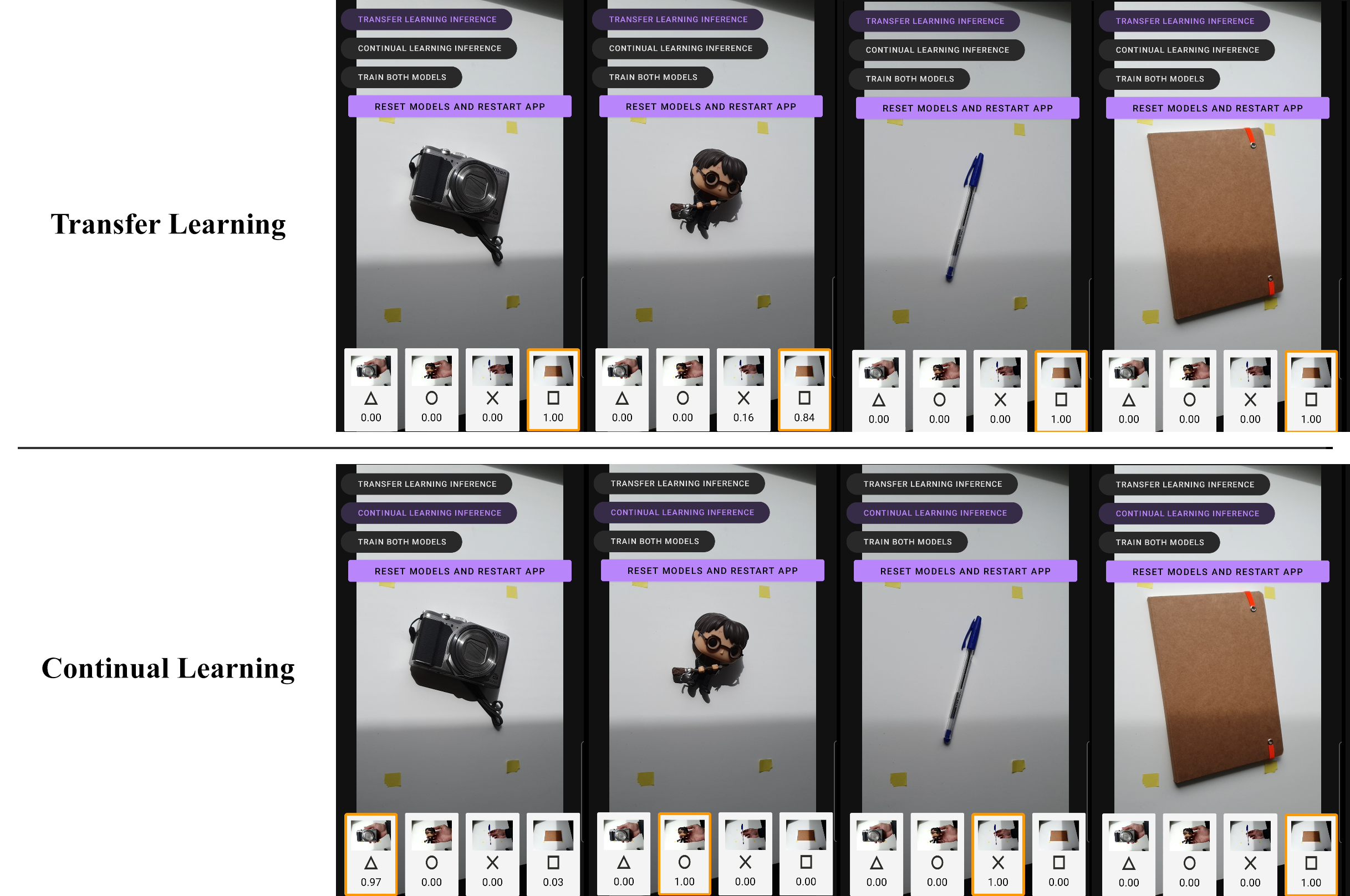}
\end{center}
   \caption{Object classification and confidence scores for both Transfer and Continual Learning models, recorded on the Android application during the New Classes Scenario experiments. Continual Learning has no issue classifying the objects. Transfer Learning fails dramatically due to catastrophic forgetting of old classes, caused by incremental training.}
\label{fig:classesscenario}
\end{figure*}

The real strength and usefulness of the CL model over the TL one can be seen clearly from the New Classes Scenario, which is depicted in Figure~\ref{fig:classesscenario}. In this scenario, the CL model correctly classifies all four classes with near-perfect scores. The TL model, however, fails dramatically and is able to correctly classify only one out of the four classes. Specifically, it only correctly classifies the notebook class, which is the last one introduced during the incremental training scenario.
This constitutes clear evidence of catastrophic forgetting when the model is trained using TL in a more realistic, class-incremental learning scenario while deployed on an embedded device.

\subsection{Testing Continual Learning Under Non-Ideal Conditions}

We have already shown that extending TF Lite capabilities to allow for continual learning models to be developed and deployed on embedded devices offers a variety of benefits when it comes to applying deep learning solutions to more realistic scenarios on the edge. The next step is to examine the robustness of the CL model under non-ideal conditions. For this reason, we train and test the model on four new classes, which can be seen in Figure~\ref{fig:extreme}. In contrast to the previous experiments, we now collect 70 unique samples for each class with varying characteristics and difficulty. Specifically, a sample might have a different orientation or depth, the background and light conditions are constantly changing, the object in the image might be cropped or blurry, and some unique characteristics of the object might be hidden. A video showing the sample collection, incremental training and testing procedure can be found at this link: \url{https://www.youtube.com/watch?v=mVI1ob55vZw&}.

When testing under such realistic conditions, the CL model still manages to correctly classify the objects in the camera feed, most of the time. Specifically, the confidence score for the Remote and Bottle classes ranges from 0.85 to 1, for the Cup class ranges from 0.75 to 1 and for the Glasses class is always 1. Despite its generally good performance, the CL model fails to correctly classify the objects in specific situations such as:
\begin{itemize}
  \item when the Remote control is rotated as to not show the buttons (key characteristic hidden)
  \item when the handle of the Cup is not visible (cropped or blurry)
  \item when only the bottom of the Bottle is shown and therefore the cap is not visible (key characteristic hidden)
\end{itemize}

It is also particularly interesting that every time the CL model fails, it always defaults to the last-learned class, which in this case is the Glasses class. This might be mitigated, in future work, by adjusting the learning rate or by mixing and shuffling the new samples with the replay buffer samples while training, instead of executing replay at the end of the training session.


\section{Conclusion and Future Work} \label{sec5}

In this paper we have shown that the current TensorFlow Lite transfer learning implementation is incapable of handling realistic, continual learning tasks, as simulated through the CORe50 benchmark and demonstrated through our own Android demo application. To solve this issue, we expanded Google’s TF Lite library to include continual learning capabilities, by introducing a simple replay buffer into the current TL model. The code for our demo application will be open-sourced with the goal of enabling developers to integrate continual learning to their smartphone applications, as well as facilitating further development of continual learning functionality into the TF Lite environment.

In the future, we will focus on implementing additional continual learning methodologies, like regularization, modularity and more sophisticated rehearsal techniques, within the limitations of TF Lite, by altering the head network of the model, and testing our models, head-to-head, with several current state-of-the art algorithms. We will experiment with adding more layers to the \emph{head} and increasing the overall complexity of the network to allow for tuning the representation as well. Finally, we intend to implement a dedicated continual learning TF API for Android to facilitate further development and experimentation in this sector.


\section*{Acknowledgement}
This project has received funding from the European Union's Horizon 2020 Research and Innovation Programme under Grant Agreement No 739578 and the Government of the Republic of Cyprus through the Deputy Ministry of Research, Innovation and Digital Policy.

{\small
\bibliographystyle{ieee_fullname}
\bibliography{egbib}

\begin{thebibliography}{10}\itemsep=-1pt

\bibitem{cortes2017adanet}
Corinna Cortes, Xavier Gonzalvo, Vitaly Kuznetsov, Mehryar Mohri, and Scott
  Yang.
\newblock Adanet: Adaptive structural learning of artificial neural networks.
\newblock In {\em International conference on machine learning}, pages
  874--883. PMLR, 2017.

\bibitem{deng2009imagenet}
Jia Deng, Wei Dong, Richard Socher, Li-Jia Li, Kai Li, and Li Fei-Fei.
\newblock Imagenet: A large-scale hierarchical image database.
\newblock In {\em 2009 IEEE conference on computer vision and pattern
  recognition}, pages 248--255. Ieee, 2009.

\bibitem{doshi2020continual}
Keval Doshi and Yasin Yilmaz.
\newblock Continual learning for anomaly detection in surveillance videos.
\newblock In {\em Proceedings of the IEEE/CVF Conference on Computer Vision and
  Pattern Recognition Workshops}, pages 254--255, 2020.

\bibitem{farquhar2018towards}
Sebastian Farquhar and Yarin Gal.
\newblock Towards robust evaluations of continual learning.
\newblock {\em arXiv preprint arXiv:1805.09733}, 2018.

\bibitem{hadsell2020embracing}
Raia Hadsell, Dushyant Rao, Andrei~A Rusu, and Razvan Pascanu.
\newblock Embracing change: Continual learning in deep neural networks.
\newblock {\em Trends in Cognitive Sciences}, 2020.

\bibitem{howard2017mobilenets}
Andrew~G Howard, Menglong Zhu, Bo Chen, Dmitry Kalenichenko, Weijun Wang,
  Tobias Weyand, Marco Andreetto, and Hartwig Adam.
\newblock Mobilenets: Efficient convolutional neural networks for mobile vision
  applications.
\newblock {\em arXiv preprint arXiv:1704.04861}, 2017.

\bibitem{hsu2018re}
Yen-Chang Hsu, Yen-Cheng Liu, Anita Ramasamy, and Zsolt Kira.
\newblock Re-evaluating continual learning scenarios: A categorization and case
  for strong baselines.
\newblock {\em arXiv preprint arXiv:1810.12488}, 2018.

\bibitem{ioffe2017batch}
Sergey Ioffe.
\newblock Batch renormalization: Towards reducing minibatch dependence in
  batch-normalized models.
\newblock {\em arXiv preprint arXiv:1702.03275}, 2017.

\bibitem{kemker2017fearnet}
Ronald Kemker and Christopher Kanan.
\newblock Fearnet: Brain-inspired model for incremental learning.
\newblock {\em arXiv preprint arXiv:1711.10563}, 2017.

\bibitem{kirkpatrick2017overcoming}
James Kirkpatrick, Razvan Pascanu, Neil Rabinowitz, Joel Veness, Guillaume
  Desjardins, Andrei~A Rusu, Kieran Milan, John Quan, Tiago Ramalho, Agnieszka
  Grabska-Barwinska, et~al.
\newblock Overcoming catastrophic forgetting in neural networks.
\newblock {\em Proceedings of the national academy of sciences},
  114(13):3521--3526, 2017.

\bibitem{lemke2015metalearning}
Christiane Lemke, Marcin Budka, and Bogdan Gabrys.
\newblock Metalearning: a survey of trends and technologies.
\newblock {\em Artificial intelligence review}, 44(1):117--130, 2015.

\bibitem{li2019rilod}
Dawei Li, Serafettin Tasci, Shalini Ghosh, Jingwen Zhu, Junting Zhang, and
  Larry Heck.
\newblock Rilod: near real-time incremental learning for object detection at
  the edge.
\newblock In {\em Proceedings of the 4th ACM/IEEE Symposium on Edge Computing},
  pages 113--126, 2019.

\bibitem{li2017learning}
Zhizhong Li and Derek Hoiem.
\newblock Learning without forgetting.
\newblock {\em IEEE transactions on pattern analysis and machine intelligence},
  40(12):2935--2947, 2017.

\bibitem{lomonaco2017core50}
Vincenzo Lomonaco and Davide Maltoni.
\newblock Core50: a new dataset and benchmark for continuous object
  recognition.
\newblock In {\em Conference on Robot Learning}, pages 17--26. PMLR, 2017.

\bibitem{lomonaco2019rehearsal}
Vincenzo Lomonaco, Davide Maltoni, and Lorenzo Pellegrini.
\newblock Rehearsal-free continual learning over small non-iid batches.
\newblock {\em arXiv preprint arXiv:1907.03799}, 2019.

\bibitem{lopez2017gradient}
David Lopez-Paz and Marc'Aurelio Ranzato.
\newblock Gradient episodic memory for continual learning.
\newblock {\em arXiv preprint arXiv:1706.08840}, 2017.

\bibitem{mccloskey1989catastrophic}
Michael McCloskey and Neal~J Cohen.
\newblock Catastrophic interference in connectionist networks: The sequential
  learning problem.
\newblock In {\em Psychology of learning and motivation}, volume~24, pages
  109--165. Elsevier, 1989.

\bibitem{pan2009survey}
Sinno~Jialin Pan and Qiang Yang.
\newblock A survey on transfer learning.
\newblock {\em IEEE Transactions on knowledge and data engineering},
  22(10):1345--1359, 2009.

\bibitem{parisi2019continual}
German~I Parisi, Ronald Kemker, Jose~L Part, Christopher Kanan, and Stefan
  Wermter.
\newblock Continual lifelong learning with neural networks: A review.
\newblock {\em Neural Networks}, 113:54--71, 2019.

\bibitem{pellegrini2019latent}
Lorenzo Pellegrini, Gabriele Graffieti, Vincenzo Lomonaco, and Davide Maltoni.
\newblock Latent replay for real-time continual learning.
\newblock {\em arXiv preprint arXiv:1912.01100}, 2019.

\bibitem{senchanka_2019}
Pavel Senchanka.
\newblock Example on-device model personalization with tensorflow lite, Dec
  2019.

\bibitem{van2019three}
Gido~M Van~de Ven and Andreas~S Tolias.
\newblock Three scenarios for continual learning.
\newblock {\em arXiv preprint arXiv:1904.07734}, 2019.

\bibitem{zhou2012online}
Guanyu Zhou, Kihyuk Sohn, and Honglak Lee.
\newblock Online incremental feature learning with denoising autoencoders.
\newblock In {\em Artificial intelligence and statistics}, pages 1453--1461.
  PMLR, 2012.

\end{thebibliography}
}

\end{document}